\documentclass[10pt,twocolumn,letterpaper]{article}

 \usepackage{cvpr}              
\usepackage{algorithm}
\usepackage[accsupp]{axessibility}
\usepackage{color}
\usepackage{algpseudocode}
\usepackage{amsmath}
\usepackage{amsthm}
\usepackage{multirow}

\usepackage[dvipsnames]{xcolor}

\definecolor{cvprblue}{rgb}{0.21,0.49,0.74}
\usepackage[pagebackref,breaklinks,colorlinks,citecolor=cvprblue]{hyperref}

\def\paperID{13292} 
\def\confName{CVPR}
\def\confYear{2024}

\title{Topology Sculptor, Shape Refiner: Discrete Diffusion Model for High-Fidelity 3D Meshes Generation}

\author{
Kaiyu Song\thanks{This work was done during the internship at Tencent.}, Hanjiang Lai\thanks{Hanjiang Lai is the corresponding author.}, Yan Pan, Jian Yin\\
Sun Yat-sen University\\
Guangdong, China\\
{\tt\small songky7@mail2.sysu.edu.cn, \{laihanj3, panyan5, issjyin\}@mail.sysu.edu.cn}
\and
Yaqing Zhang, Chuangjian Cai\\
Tencent VisVise\\
Guangdong, China\\
{\tt\small \{yaqingzhang, herbertcai\}@tencent.com}
\and
Kun Yue\\
Yunnan University\\
Yunnan, China\\
{\tt\small kyue@ynu.edu.cn}
}

\begin{document}
\maketitle
\begin{abstract}

In this paper, we introduce Topology Sculptor, Shape Refiner (TSSR), a novel method for generating high-quality, artist-style 3D meshes based on Discrete Diffusion Models (DDMs). Our primary motivation for TSSR is to achieve highly accurate token prediction while enabling parallel generation, a significant advantage over sequential autoregressive methods. By allowing TSSR to "see" all mesh tokens concurrently, we unlock a new level of efficiency and control. We leverage this parallel generation capability through three key innovations: 1) Decoupled Training and Hybrid Inference, which distinctly separates the DDM-based generation into a topology sculpting stage and a subsequent shape refinement stage. This strategic decoupling enables TSSR to capture both intricate local topology and overarching global shape effectively. 2) An Improved Hourglass Architecture, featuring bidirectional attention enriched by face-vertex-sequence level Rotational Positional Embeddings (RoPE), thereby capturing richer contextual information across the mesh structure. 3) A novel Connection Loss, which acts as a topological constraint to further enhance the realism and fidelity of the generated meshes. Extensive experiments on complex datasets demonstrate that TSSR generates high-quality 3D artist-style meshes, capable of achieving up to 10,000 faces at a remarkable spatial resolution of $1024^3$. The code will be released at: https://github.com/psky1111/Tencent-TSSR.

\end{abstract}    
\section{Introduction}

High-fidelity 3D mesh generation~\cite{hunyuan2.0,meshanythingv2} has garnered significant attention, focusing on generating the desired shape and high-quality topology. It produces detailed, topologically sound geometric representations for a wide range of applications, including virtual reality, digital content creation, and robotic simulation.

Mainstream 3D mesh generation methods can be broadly categorized into two approaches: 1) \textit{Shape-then-mesh generation} and 2) \textit{direct mesh generation}. Shape-then-mesh generations~\cite{hunyuan2.0} start with representing a shape via continuous volumetric fields, such as Neural Radiance Fields (NeRF)~\cite{NeRF} and Signed Distance Fields (SDF)~\cite{hunyuan2.1}. Then, a diffusion model is used to generate such a representation and convert it to real meshes via Marching Cubes~\cite{marchingcubes}. Benefits from parallel token generation in diffusion models, but shape-then-mesh generation is efficient, but suffers from inevitable denegation after the conversion process. Direct mesh generation methods aim to predict mesh vertices and faces directly. Mainstream methods, such as Meshtron~\cite{meshtron} and DeepMesh~\cite{deepmesh}, employ the Auto-Regressive (AR) approach, which first quantizes the mesh coordinates into a discrete space (V discrete bins for each coordinate, where V is also referred to as the spatial resolution in this context). Then it implements the specific tokenizer method to generate the mesh token by token. Benefits from the discrete space, ARs can generate high-quality meshes, but struggle with efficiency.

Discrete Diffusion Models (DDMs) serve as a remedy to enable parallel generation in a discrete space, such as SeedDiffusion~\cite{seeddiffusion} and LLaDA \cite{llada} in large language models (LLMs). DDMs follow a similar process to the diffusion models, which contain: 1) the forward process, adding noise to the discrete token sequence, and 2) the reverse process, denoising the noise-corrupted token sequence. However, token-parallel generation inherently breaks the causal sequence (the generated order), leading to less accurate token prediction when encountering long, high-resolution sequences. In 3D mesh generation, this problem is vastly magnified, as the causal sequence can directly ensure valid topology and precise token prediction~\cite{meshtron,meshanythingv2}. A natural question thus arises: "How to maintain accurate token prediction in long and high spatial sequences when using parallel generation (which breaks causal order)?"

In this paper, we propose the Topology Sculptor, Shape Refiner (TSSR), a novel paradigm explicitly designed for direct mesh generation built upon DDMs. TSSR addresses the challenge of maintaining accurate token prediction in parallel generation for meshes. TSSR leverages DDMs' inherent ability to generate tokens in parallel, enabling the model to view all tokens simultaneously and to utilize bidirectional attention to capture comprehensive global shape and intricate topological information. TSSR realizes accurate parallel mesh generation through three integrated core contributions:

1) A unique decoupling training and hybrid inference paradigm that strategically divides mesh generation into specialized sub-tasks for shape (shape refiner) and topology (topology sculptor). This strategy acknowledges that while parallel generation yields rich, comprehensive information (encompassing global shape and intricate topology), effectively processing this intertwined complexity for accurate prediction is challenging. Decoupling training enables separating these aspects into distinct sub-tasks, thereby reducing the learning burden on the model and allowing each sub-task to focus predominantly on either shape or topology. Then, the hybrid inference process unifies these subtasks into a complete, iterative refinement loop, enabling mutually enhancing shape and topology prediction and thereby improving token prediction accuracy. 2) An improved Hourglass architecture designed to process and leverage multi-level mesh information effectively. This architecture integrates Multi-level RoPE and an enhanced transformer block to capture crucial hierarchical geometric context, enabling highly accurate token prediction across long, spatially complex sequences. 3) A novel connection loss that directly introduces topology priors during training. This loss enforces hard topology constraints by ensuring token consistency across vertices, thereby significantly enhancing the model’s ability to maintain topological integrity. Collectively, these three components ensure that TSSR fully utilizes the rich information available in parallel token generation. This strategic integration significantly improves token prediction accuracy, thereby addressing the loss of sequential causal dependencies and rendering parallel mesh generation highly effective for artist-style meshes. Experimental results on various open datasets demonstrate TSSR's ability to generate artist-style meshes via parallel token generation at high resolutions (i.e., $1024^{3}$) and up to 10K faces.

To summarize, our key contributions are as follows:
\begin{itemize}
    \item We propose the Topology Sculptor, Shape Refiner, a novel DDM-based paradigm specifically tailored for artist-style 3D mesh generation. 
    
    \item By carefully designing, TSSR enables full leverage of the rich information after "seeing" all tokens in token-parallel generation.
    
    \item We demonstrate that TSSR achieves state-of-the-art performance in generating high-quality 3D artist-style meshes with up to 10,000 faces at a spatial resolution of $1024^3$, showcasing superior global coherence and local precision.
\end{itemize}

\section{Related Work}

\textbf{Shape-then-mesh generation.} These works focus on first generating the shape via a continuous volumetric field, where SDF is the most common choice. Then, SDF can be converted to a mesh using the Marching Cubes~\cite{marchingcubes}. For example, TRELLIS~\cite{TRELLIS} and Hunyuan 2.5~\cite{hunyuan2.5} first leverage the VAE to reconstruct the mesh via SDF. Then, they leverage the diffusion model to generate the VAE latent. The advantage is that the explicit 3D information can be incorporated into the VAE. Diffusion itself can be more robust to the 3D shape. The difference is that TRELLIS introduces the explicit shape constraint into VAE training. Hunyuan proposed an improved version for extracting GT SDF. Besides enhancing VAE, Wonder3D~\cite{wonder3d} introduces a cross-domain diffusion model framework to further enhance the diffusion process. 

\textbf{Direct mesh generation.} Direct mesh generation aims at generating a mesh without conversion, such as SDF to mesh. The mainstream way is the ARs method. For example, MeshGPT~\cite{MeshGPT}  pioneers AR-based mesh synthesis through sequence-based modeling. MeshAnything~\cite{meshanything} and MeshAnything V2~\cite{meshanythingv2} proposed an improved tokenizer through adjacent mesh tokenization. MeshXL~\cite{MeshXL} introduced the Neural Coordinate Field, which fuses explicit coordinate
representation with implicit neural embeddings to enhance the AR. EdgeRunner~\cite{edgerunner} proposed compressing variable-length meshes into fixed-size latent vectors to reduce the sequence length, which improves the efficiency of ARs. Meshtron~\cite{meshtron} proposed the Hourglass architecture to enhance mesh generation quality and efficiency further. MeshMosaic~\cite{MeshMosaic} proposed the part-splitting-based method to allow parallel generation of the part. 

Except for the ARs, there is some exploration of token-parallel generation via diffusion. For example, Meshcraft~\cite{meshcraft} and Polydiff~\cite{polydiff} proposed directly mapping the mesh into a continuous latent space via VAE and then implementing diffusion models on the latent space. However, the compression loss in VAE will inevitably lead to further degradation.

\textbf{Discrete Diffusion Models.} DDMs~\cite{dfm,fmgdp} are another type of diffusion model compared to the continuous state-based diffusion models~\cite{fm}. Previous work has begun to implement DDMs across various tasks, such as LLMs~\cite {seeddiffusion} and image generation~\cite{FUDOKI}, demonstrating their potential. In this work, we propose TSSR, a token parallel generation paradigm based on the DDM. TSSR directly generates artist-like methods without any conversion, such as SDF and VAE, thereby maintaining high quality.

\section{Preliminaries}
This section provides essential background on discrete mesh representation and the general framework of DDMs as applied to mesh generation.

\textbf{Discrete representation of mesh.} A 3D mesh $M$ is fundamentally represented by a set of $N$ faces, denoted as $\{\text{f}_{i}\}_{i=1}^N$. Each face $\text{f}_i$ is composed of $n$ vertices, $\{\text{v}_{i,j}\}_{j=1}^n$. In this paper, we specifically consider triangular meshes, thus setting $n=3$, where each face has three vertices. Each vertex $\text{v}_{i,j}$ is defined by its three-dimensional Cartesian coordinates: $\{v_{i,j}x, v_{i,j}y, v_{i,j}z\}$. To enable processing by discrete generative models, vertex coordinates are first discretized into a fixed number of bins, defining the spatial resolution. Consistent with the high-fidelity demands of artist-style mesh generation, we adopt a spatial resolution of $1024^3$, meaning coordinates are quantized into $V=1024$ discrete bins along each axis.
For discrete diffusion models on meshes, these discretized coordinates are transformed into a sequence of tokens. We utilize the Meshtron tokenizer \cite{meshtron}, which flattens all vertex coordinates of a mesh into a one-dimensional sequence. Specifically, a tokenized mesh $\hat{x}_{1}$ is represented as:
\begin{equation}
\hat{x}_{1} = \{v_{1,1}x, v_{1,1}y,v_{1,1}z, \ldots, v_{N,n}x, v_{N,n}y,v_{N,n}z\}.
\label{eq:tokenizer}
\end{equation}
The sequence length of $\hat{x}_{1}$ is $9N$, as each of the $N$ triangular faces contributes 9 coordinate tokens (3 vertices $\times$ 3 coordinates per vertex) \cite{meshtron}. The effective codebook size for these tokens is $B=V+k$, where $V=1024$ represents the quantized coordinate values and $k$ denotes the number of any additional special tokens (e.g., start-of-sequence, end-of-sequence, padding) used in the tokenization scheme.

\textbf{Discrete Diffusion Models (DDMs)}
Discrete Diffusion Models (DDMs) are a class of generative paradigms that learn to transform pure noise into target data via discrete space. This is achieved by defining a forward diffusion process that gradually corrupts clean data ($\hat{x}_1$) into noise ($x_0$), and then learning to reverse this process to generate data. Unlike continuous diffusion models that often define a velocity field in a continuous space, DDMs are characterized by a model that predicts the clean target state, $p_{1|t}(x_t, t, c, \theta)$, from a noisy state $x_t$ at time $t \in [0,1]$ (where $t=0$ represents initial noise and $t=1$ represents the clean state).

The core of DDM inference lies in defining an iterative update rule that transitions from a current noisy state $x_t$ to a slightly cleaner state $x_{t+\Delta t}$ (or $x_{t-\Delta t}$), leveraging the predicted clean state $p_{1|t}$. Such a transition could be achieved by two views:\textit{Probability Space (e.g., Discrete Flow Matching \cite{dfm}).} $p_{1|t}$ is used to define a continuous-time velocity field in a probability space, which then guides the transition from $x_t$ to $x_{t+\Delta t}$. This often involves complex mathematical derivations for the update rule.

\textit{Token Space Updates (e.g., as in SeedDiffusion \cite{seeddiffusion} and TSSR ):} In this approach, $p_{1|t}(x_t, t, c, \theta)$ is directly interpreted as a categorical distribution over tokens. The update rule for $x_t \to x_{t+\Delta t}$ is then constructed by sampling from this $p_{1|t}$ prediction to decide which clean tokens to incorporate, or which tokens to mask/randomize, to form the next state $x_{t+\Delta t}$. This method operates directly in the discrete token space, making it more intuitive.

In DDMs, two primary types of noise are commonly used to define the corruption process in the forward pass:
\begin{enumerate}
\item \textbf{Mask-based Noise:} Corrupts tokens by replacing ground truth values with a special mask, typically denoted as $[\text{MASK}]$. This formulation inherently supports inpainting tasks, as the model learns to fill in missing information based on surrounding context.
\item \textbf{Uniform-based Noise:} Corrupts tokens by replacing ground truth values with random tokens sampled uniformly from the entire valid codebook. This approach encourages the model to learn global structural recovery, as it must reconstruct the data from entirely random inputs.
\end{enumerate}

The forward process of a DDM defines the distribution of a noisy state $x_t$ at time $t \in [0,1]$, given the clean data $\hat{x}_1$ and the related noise data $x_{0}$:
\begin{equation}
p(x_{t} | \hat{x}_{1},x_{0}) = \kappa_t \delta_{x_{0}}(x_{0}) + (1 - \kappa_t) \delta_{\hat{x}_1}(\hat{x}_1),
\label{eq:forward_process_ddm}
\end{equation}
where $\kappa(t)$ is a time-dependent noise scheduler that controls how many tokens in $\hat{x}_{1}$ will be replaced by noise tokens $x_{0}$. $\delta_{x}(y)$ is a Dirac delta function (or its discrete equivalent, a one-hot distribution) indicating that $x = y$.

The inference process in DDMs aims to generate samples from $p(\hat{x}_1)$ by starting from a noisy state (e.g., pure mask tokens or random tokens) and iteratively refining it from $t=0$ (noise) to $t=1$ (clean) or vice versa. The model predicts the clean target state, $p_{1|t}(x_{t},t,c,\theta)$, and this prediction is then used by a custom update rule to transition to the next state in the refinement sequence.

The training process for DDMs optimizes the model parameters $\theta$ to predict the final clean state. This is commonly achieved by minimizing a cross-entropy loss between the model's prediction of the clean state and the actual clean state $\hat{x}_1$:
\begin{equation}
\mathcal{L}_{DDM} = \text{CE}(p_{1|t}(x_{t},t,c,\theta),\hat{x}_{1}),
\label{eq:loss_ddm}
\end{equation}
where $x_t$ is sampled from the forward process given $\hat{x}_1$.

\section{Method} 
\label{sec:method}

\textbf{Problem definition.} Our target is to generate artist-like meshes given the point cloud condition $c$ via DDMs to enable parallel token generation. The overall pipeline contains inference and training. During training, we can access GT meshes and related $c$. We convert GT meshes into quantized, discrete token sequences $\hat{x}_{1}$ using the tokenizer. Then, we add noise into the $\hat{x}_{1}$ to get $x_{t}$. In the end, we train a model to predict the clean token sequence. During inference, we cannot access GT meshes. We will initialize with arbitrary token sequences composed of pure noise tokens. Then, we iteratively denoise the noise via a trained model. In the end, we obtain a final clean token sequence $x_{1}$ and a de-tokenizer to generate meshes. 

In this paper, TSSR effectively leverages the global shape and intricate topology available in parallel token generation to propose novel training and inference via three integrated core contributions: 1) a unique decoupling training and hybrid inference, 2) an improved Hourglass architecture, and 3) the proposed connection loss.

\textbf{Decoupling via inherent noise biases.} The TSSR paradigm is motivated by the distinct learning biases inherent in the two primary DDMs noise types, which we leverage to decouple the complex task of mesh generation.
\begin{itemize}
\item Mask-based noise for topology sculpting: As an infilling task, denoising mask-corrupted data naturally drives the model to focus on local structural completion, such as maintaining connectivity and filling small holes. This inherent behavior biases the model to act as a powerful \textit{topology sculptor}.

\item Uniform-based noise for shape refining: Denoising from a "totally broken shape" of random valid tokens forces the model to recover the overall global form and proportions, as local structural priors are destroyed. This naturally biases the model to function as a powerful \textit{shape refiner}.
\end{itemize}
\begin{figure}
    \centering
    \includegraphics[width=0.5\linewidth]{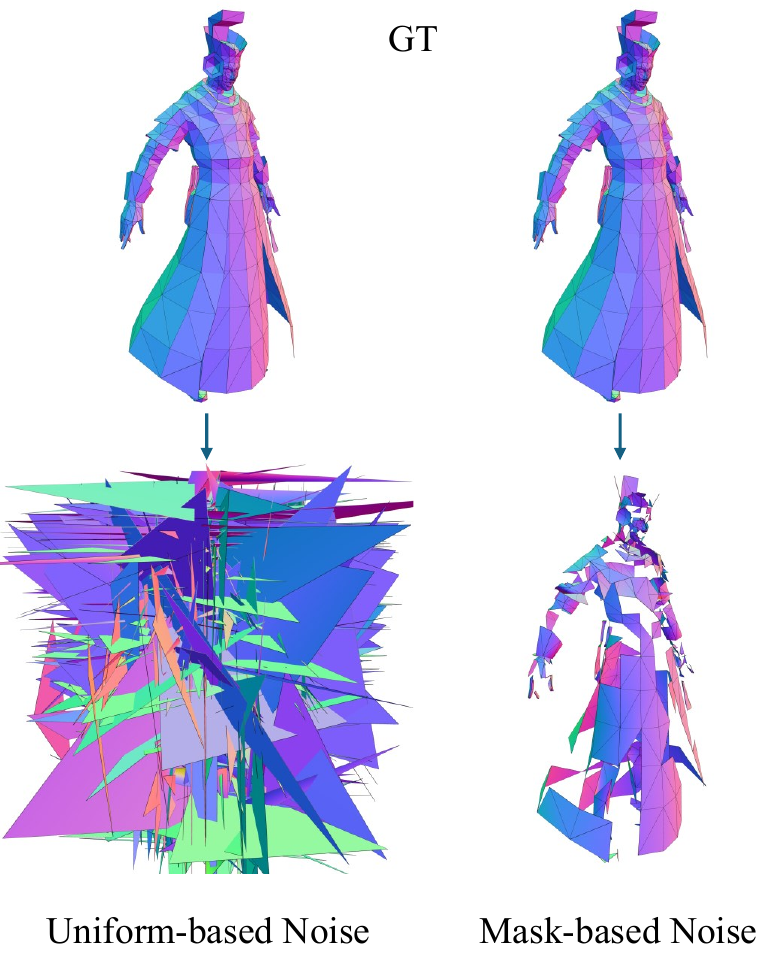}
    \caption{An illustration of different types of noise for meshes. Then [MASK] is treated as a token removal. As a result, mask-based noise will force the model to focus on local structural completion. Uniform noise will generate a "hedgehog"-like mesh, leading the model to focus on the shape.}
    \label{fig:motivation}
\end{figure}
We visualize the corrupted mesh via two noises, as shown in Fig.~\ref{fig:motivation}, to further illustrate the learning biases. Leveraging this insight, TSSR develops a hybrid process that combines these two specialized roles to achieve both accurate topology and coherent shape.

\textbf{Hybrid inference and decoupling training paradigm.} Firstly, we demonstrate the hybrid inference process to clearly show how TSSR transfers the state from $x_{t}$ to $x_{t+\Delta t}$. We assume there are two pre-trained models, $\theta_{mask}$ and $\theta_{uniform}$, which are responses to the topology sculptor and the shape refiner, respectively. Then, the inference process involves: 1) the initial states, which decide what type of noise in $x_{t}$ and $x_{t + \Delta t}$, and 2) the iterative update rule from $x_{t}$ to $x_{t+\Delta t}$. The inference process starts with the pure mask token-based sequence, i.e., $x_{0}$ is composed of [MASK]. This is the trick from the theoretical gargantuan to enable customization of the overall inference process and to ignore the rates of GT and noise tokens during training~\cite {dfm}. Therefore, $x_{t}$ and $x_{t + \Delta t}$ are the masked-noise-corrupted versions.

The iterative rule involves cycling through the topology sculptor $\theta_{mask}$ and $\theta_{uniform}$, as well as a classifier $\phi$, to decide which tokens to keep. We define $x_{t, mask}$ and $x_{t, uniform}$ to represent the mask-based noise corrupted and uniform-based noise corrupted $x_{t}$, respectively. Then, $x_{t}$, $t<1$ , always have mask noise and thus $x_{t,massk} = x_{t}$. We do the mask-based denoising process to calculate $p_{1|t}(x_{t,mask},t,c,\theta_{mask})$ and get its related clean state prediction $x_{1|t,mask}$. $x_{1|t,mask}$ will inevitably have some errors in token results due to the bias of $\theta_{mask}$, but keep some local intricate topology based on our motivation. Thus, it can naturally be regarded as adding a special type of uniform-based noise due to model bias. This allows TSSR to use the shape refiner to correct these error tokens again and recover the shape, where we have $x_{t,uniform} = x_{1|t,mask}$. We then do uniform-based denoising process to calculate $p_{1|t}(x_{t,uniform},t,c,\theta_{uniform})$ and get its related clean state prediction $x_{1|t,uniform}$. 

Now, the final step is to decide how many tokens to keep and transfer the rest to [MASK] to obtain $x_{t + \Delta t}$. We leverage $\phi$ to calculate:
\begin{equation}
     \text{Softmax}(p_{\phi}(x_{1|t,uniform})) < \sigma,
     \label{eq:mask_rate}
\end{equation}
where $\text{Softmax}(\ast)$ is the softmax operator, $p_{\phi}(x_{1|t,uniform})$ is the logits about whether the tokens are correct via classifier $\phi$. Eq.~\ref{eq:mask_rate} defines the confidence about the $x_{1|t,uniform}$. For each token in $x_{1|t,uniform}$, the confidence below than $\sigma$ will be replaced by [MASK]. In this way, we can $x_{t + \Delta t}$ after re-masking by Eq.~\ref{eq:mask_rate} and finish one iteration via our hybrid inference.

Then, we demonstrate our decoupling training. Hybrid inference are related to three models $\theta_{mask}$, $\theta_{uniform}$ and $\phi$. For $\theta_{mask}$, it follows the standard mask noise-based DDMs. We generate $x_{t, mask}$ via Eq.~\ref{eq:forward_process_ddm}. Then, the $\theta_{mask}$ is to predicted the clean state $p_{1|t}(x_{t,mask},t,c,\theta_{mask})$, where the loss can be formulated via Eq.~\ref{eq:loss_ddm} as:
\begin{equation}
     \mathcal{L}_{mask} = \text{CE}(p_{1|t}(x_{t},t,c,\theta_{mask}),\hat{x}_{1}).
     \label{eq:mask_loss}
\end{equation}

Then, for $\theta_{uniform}$, we follow the hybrid inference to calculate $x_{t, uniform}$ via $x_{t,uniform} \sim p_{1|t}(x_{t,mask},t,c,\theta_{mask})$ and the loss can be formulated:
\begin{equation}
         \mathcal{L}_{uniform} = \text{CE}(p_{1|t}(x_{t},t,c,\theta_{uniform}),\hat{x}_{1}).
         \label{eq:uniform_loss}
\end{equation}
Similarly, we directly calculate $p_{\phi}(x_{1|t,uniform})$ via $x_{1|t,uniform} \sim p_{1|t}(x_{t},t,c,\theta_{uniform})$ and build a binary classification task via:
\begin{equation}
    \mathcal{L}_{\phi} = \text{BE}(p_{\phi}(x_{1|t}),\hat{x}_{1}),
    \label{eq:classifier_loss}
\end{equation}
where $\text{BE}(\ast)$ is the binary classification loss. 

In this way, the overall loss can be defined as follows:
\begin{equation}
    \mathcal{L} = \mathcal{L}_{mask} + \mathcal{L}_{uniform} + \mathcal{L}_{\phi}
\end{equation}

\begin{figure*}
    \centering
    \includegraphics[width=0.9\linewidth]{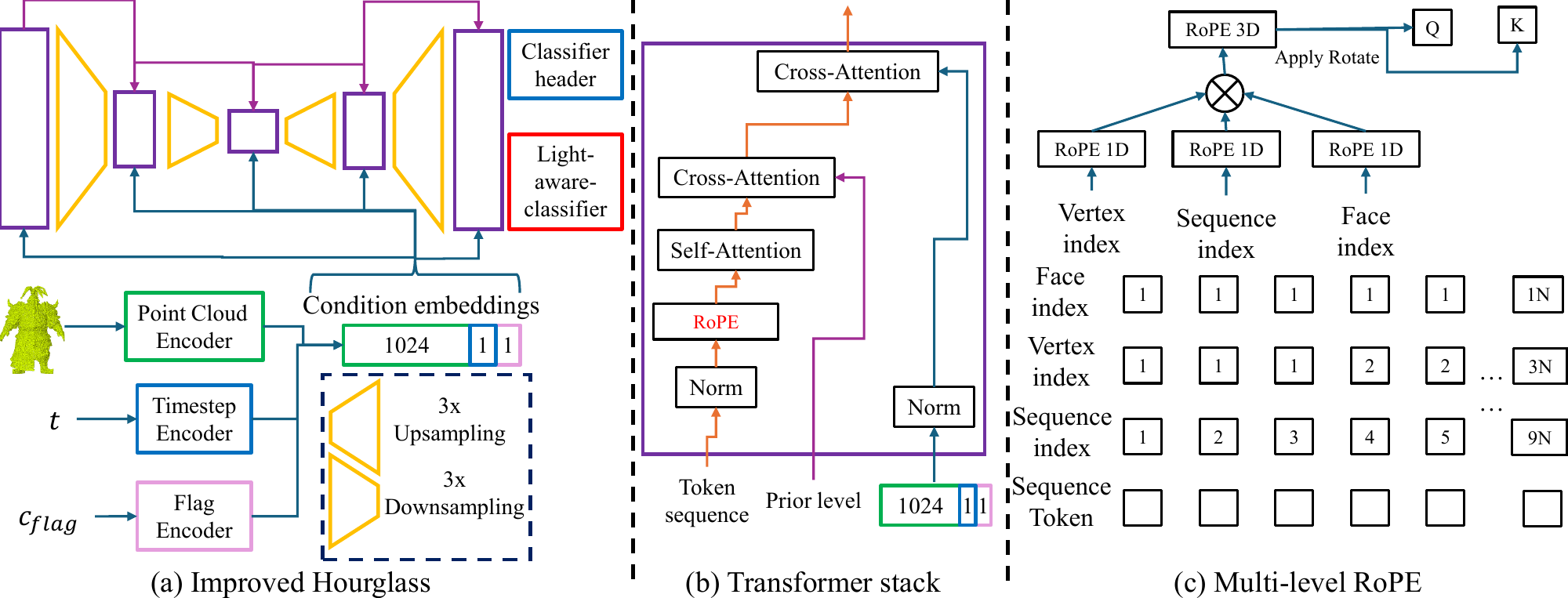}
    \caption{An illustration of improved Hourglass architecture. (a) illustrates the overall framework, where we add $c_{flag}$, an additional light-aware classifier to represent $\phi$. The classifier heads serve as initial heads for predicting tokens. (b) illustrates that we change the initial RoPE to a multi-level RoPE, and (c) illustrates the multi-level RoPE.}
    \label{fig:hourglass_improvements}
\end{figure*}
\textbf{Improved Hourglass framework.} We improve the Hourglass framework via leveraging multi-level information (face-vertex-sequence). Meanwhile. we will show a trick to unify three models $\theta_{mask}$, $\theta_{uniform}$ and $\phi$ into one model $\theta$ with a light-aware classifier head in $\theta$ (Also noted as $\phi$ to keep consistency). The overall framework of the improved Hourglass framework is shown in Fig.~\ref{fig:hourglass_improvements}. 

Firstly, multi-level information exists in two places: 1) the Hourglass block, since Hourglass will compress the token sequence into sequence-vertex-face level first, and then reconstruction from face-vertex-sequence again. 2) The token position itself. For the Hourglass block, we add extra cross-attention and use the uncompressed level as the condition. This can help the model leverage the previous level to compress and reconstruct, potentially breaking the information bottleneck. 

Then, for the token position, we proposed Multi-level RoPE. The non-AR nature of DDMs allows every token to attend to every other token, creating an opportunity to inject richer structural information than is possible in AR models. Standard positional embeddings, such as 1D RoPE, treat the mesh sequence as a flat list, ignoring its inherent hierarchical structure (coordinates within vertices, vertices within faces). To capitalize on DDM's full attention scope, we propose a novel multi-level RoPE that explicitly encodes the mesh's hierarchical structure, shown in Fig.~\ref{fig:hourglass_improvements} (c). Instead of a single position index, each token is assigned a 3D-like positional tuple: $(\text{face\_index}, \text{vertex\_index}, \text{coordinate\_index})$. This allows the model's attention mechanism to learn relationships not only between adjacent tokens in the sequence, but also between vertices within the same face and across different faces, providing a powerful inductive bias for learning topology.

In the end, we show a trick to unify three models into one with a light-aware classifier to leverage the latent from the transformer. We first added a new condition, the conditional task flag ($c_{flag}$). The flag indicates the current task being performed (e.g., $c_{flag}=0$ for mask-based denoising, $c_{flag}=1$ for uniform-based denoising. This approach promotes parameter efficiency, allowing the model to learn shared representations of mesh structure that benefit both topology sculpting and shape refining tasks. This enables us to use the single model $\theta$ to capture the abilities of $\theta_{mask}$ and $\theta_{uniform}$. Then, we integrate a light-aware classifier $\phi$ directly into the architecture. A lightweight classification head is attached to the final layer of the Hourglass transformer. This head takes the model's final token representations as input and predicts the probability that each token belongs to a topologically incorrect region. This shared representation ensures that the features learned for generation are effectively reused for error detection.

\textbf{Connection loss.} After enabling the token parallel generation, we find that topology can be represented by correctly repeating the token among shared vertices. Since each token represents a discrete XYZ token, shared vertices will have the same token. Therefore, we leverage this to propose using connection loss as a hard constraint to further improve token prediction accuracy.

Firstly, we define the set of all valid shared vertex groups in a batch $b$ as $\mathcal{G}_b = \{G_{b,g}\}_{g=1}^{N_{groups,b}}$, where $G_{b,g}$ contains the starting token indices $s_k$ for all occurrences of the $g$-th shared vertex. For a token sequence of length $S$ and vocabulary size $V$, let $P_{b,s} \in \mathbb{R}^V$ denote the predicted logits for token $s$ in batch $b$.

The connection loss $\mathcal{L}_{connection}$ encourages a consensus in predicted logits for each coordinate dimension (X, Y, Z) across all tokenized occurrences of the same vertex. For each shared vertex group $G_{b,g} \in \mathcal{G}_b$, and for each coordinate dimension $j \in \{0, 1, 2\}$ (corresponding to X, Y, Z), we extract the predicted logits $L_{b,g,j} \in \mathbb{R}^{|G_{b,g}| \times V}$ for all its occurrences:
\begin{equation}
    L_{b,g,j} = \{ P_{b, (s_k+j)} \mid s_k \in G_{b,g}\}
\end{equation}
Then, For each $L_{b,g,j}$, the consensus logits $C_{b,g,j} \in \mathbb{R}^{V}$ are computed as the mean of its members:
\begin{equation}
    C_{b,g,j} = \frac{1}{|G_{b,g}|} \sum_{s_k \in G_{b,g}} P_{b, (s_k+j)}
\end{equation}
The consensus loss for a single group $G_{b,g}$ is the average L1 distance of individual logits from their dimension-wise consensus, averaged across the three dimensions:
\begin{equation}
    \mathcal{L}_{group}(G_{b,g}) = \frac{1}{3} \sum_{j=0}^{2}  \frac{1}{|G_{b,g}|} \sum_{s_k \in G_{b,g}} \left\| P_{b, (s_k+j)} - C_{b,g,j} \right\|_1 
\end{equation}
The final $\mathcal{L}_{connection}$ is the average of group-level losses across all valid groups in the batch:
\begin{equation}
    \mathcal{L}_{connection} = \frac{\sum_{b=1}^{B} \sum_{g=1}^{N_{groups,b}} \mathcal{L}_{group}(G_{b,g})}{\sum_{b=1}^{B} N_{groups,b}} 
    \label{eq:connect_loss}
\end{equation}
This loss directly penalizes topological errors, such as non-manifold edges or disconnected vertices, by ensuring that all predicted coordinate values for a single vertex converge to a consistent value, thereby bootstrapping topological integrity.

In this way, the final loss could be calculated as:
\begin{equation}
    \mathcal{L} = \mathcal{L}_{mask} + \mathcal{L}_{uniform} + \mathcal{L}_{\phi} + \mathcal{L}_{connection}.
\end{equation}

We summarize our hybrid inference and decoupled training via an improved Hourglass in Algorithms~\ref {al:inference} and~\ref {al:train}. In this way, TSSR successfully maintains accurate token prediction via parallel token generation. Our design enables TSSR to generate high-resolution meshes with up to 10,000 faces.

\begin{algorithm}[t]
    \caption{Decoupling training} \label{al:train}
    \begin{algorithmic}[1]
     \Statex \textbf{Input:}  Epochs
     \Statex \textbf{Output:} $\theta$,$\phi$
     \State Initialize $\theta$ and $\phi$
     \For{step in $[0,...,\text{Epochs}]$}
        \State Sample $\hat{x}_{1}$ and $c$ from Dataset.
        \State $t \sim [0,1000]$.
        \State Calculate $x_{t,mask}$ via Eq.~\ref{eq:forward_process_ddm}.
        \State Calculate $p_{1|t}(x_{t,mask},t,c,c_{flag},\theta)$
        \State Calculate $\mathcal{L}_{mask}$ via Eq.~\ref{eq:mask_loss}.
        \State $x_{t,uniform}\sim p_{1|t}(x_{t,mask},t,c,c_{flag},\theta)$
        \State Calculate $p_{1|t}(x_{t,uniform},t,c,c_{flag}$
        \State Calculate $\mathcal{L}_{phi}$ via Eq.~\ref{eq:classifier_loss}.
        \State Calculate $L_{connection}$ via Eq.~\ref{eq:connect_loss}.
        \State $\theta,\phi \gets \nabla_{\theta,\phi}\mathcal{L}$
     \EndFor
     \Statex\textbf{Return:} $\theta$,$\phi$
    \end{algorithmic}
\end{algorithm}

\begin{algorithm}[t]
    \caption{Hybrid inference} \label{al:inference}
    \begin{algorithmic}[1]
     \Statex \textbf{Input:} $\theta$,$\phi$,$\sigma$, $T$, $F$
     \Statex \textbf{Output:} $x_{1}$
     \State Initialize $x_{0}$ by Eq.~\ref{eq:tokenizer} via mask token.
     \State $x_{0,mask} \gets x_{0}$.
     \For{$t$ in $[0,...\frac{t}{T},\frac{t+\Delta t}{T},...,1]$}
        \State Calculate $p_{1|t}(x_{t,mask},t,c,c_{flag}=1,\theta)$.
        \State $x_{1|t,mask}\sim p_{1|t}(x_{t,mask},t,c,c_{flag}=1,\theta)$.
        \State $x_{t,uniform} \gets x_{1|t,mask}$.
        \State Calculate $p_{1|t}(x_{t,uniform},t,c,c_{flag}=0,\theta)$.
        \State  $x_{1|t,uniform}\sim p_{1|t}(x_{t,uniform},t,c,c_{flag}=0,\theta)$.
        \If{$t=1$}
            \State $x_{1} \gets x_{1|t,uniform}$.
        \Else
            \State Calculate $x_{t+\Delta t} $ via re-masking. \Comment{Eq.~\ref{eq:mask_rate}}
        \EndIf
     \EndFor
     \Statex\textbf{Return:} $x_{1}$
    \end{algorithmic}
\end{algorithm}

\section{Experiment}
\subsection{Implementation}
\begin{figure*}
    \centering
    \includegraphics[width=0.9\linewidth]{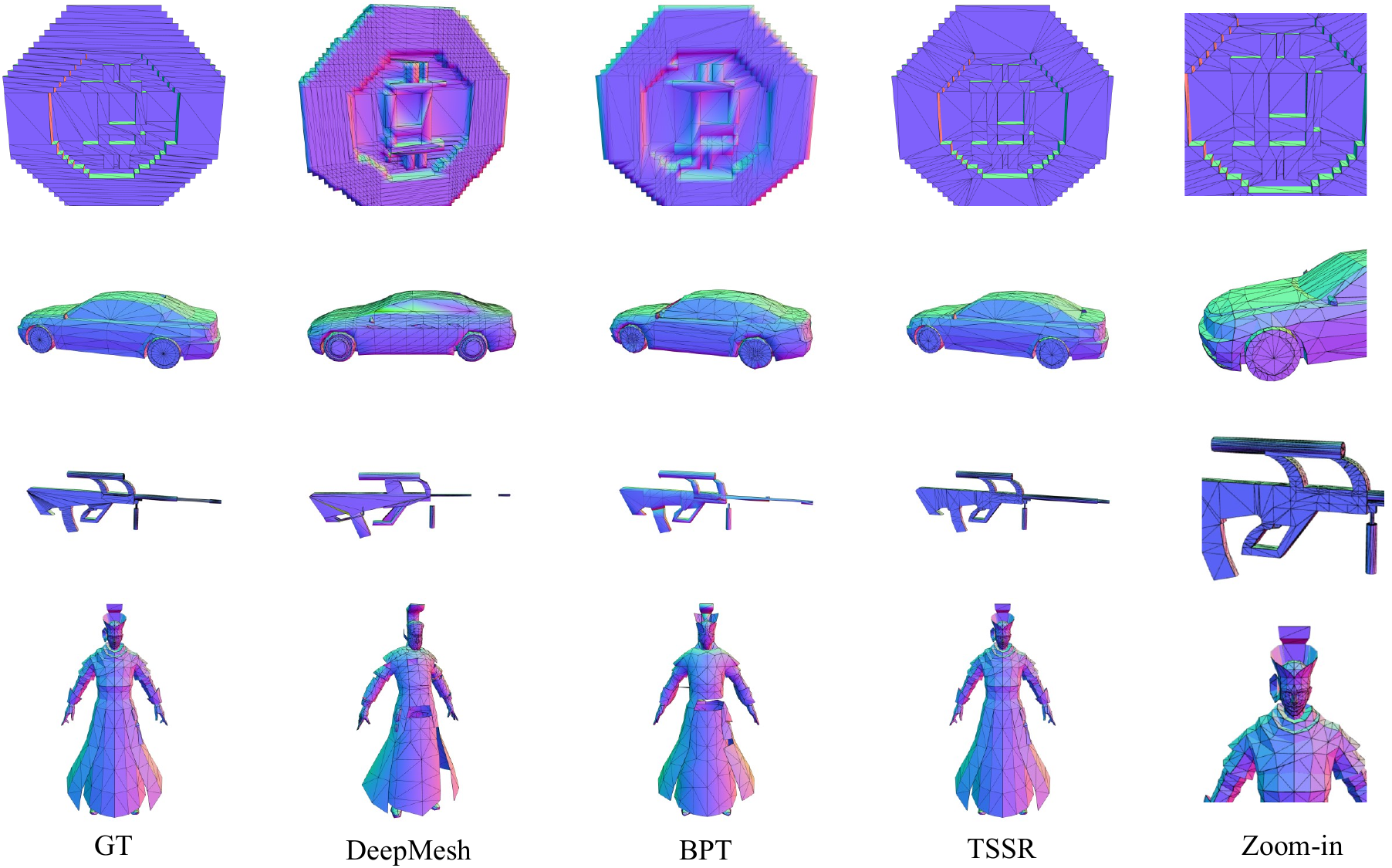}
    \caption{Qualitative results about the generated mesh. It can be found that the shape of the AR-based method will lose some parts. TSSR can benefit greatly from token parallel generation. Meanwhile, the topology of TSSR maintains a similar quality to that of AR's methods.}
    \label{fig:hourglass_improvements}
\end{figure*}
\textbf{Dataset.} The dataset consists of 400K meshes, which were collected from Objaverse-XL~\cite{objxl}, ShapeNet~\cite{shapenet}, and other licensed datasets. 

\textbf{Setting.} We train a 2B model from scratch. Training is conducted over 20 days on a cluster of 32 GPUs with 40GB MEM, initially focusing on meshes with face counts in the range $[0, 7000]$ to obtain an initial checkpoint. Then, it scales to $[7000, 10000]$ face counts trained on the 7 days using a cluster of 16 GPUs with 96GB MEM. We use a linear learning rate scheduler with a warm-up strategy and the AdamW optimizer~\cite{adamw}, increasing from $[0, 1 \times 10^{-4}]$ over the first 100 steps and maintaining $1 \times 10^{-4}$ thereafter. We set $T=200$ during inference.

\textbf{Evaluation metrics}. Following previous works~\cite{MeshMosaic}, we evaluate the quality and fidelity of the mesh using the Hausdorff Distance (HD), Chamfer Distance (CD), Normal Consistency (NC), and F-score (F1).

\textbf{Baseline.} We directly compare our method with the latest ARs method since ARs currently largely outperforms the diffusion models~\cite{MeshMosaic}. The baselines include: MeshAnythingV2, BPT~\cite{bpt}, TreeMeshGPT~\cite{TreeMeshGPT}, and DeepMesh~\cite{deepmesh}. We exclude the splitting part-like methods, such as MeshMosaic~\cite{MeshMosaic}, to ensure a fair comparison since the splitting part can cause better topology~\cite{TRELLIS}.

\subsection{Experimental Results}
\begin{table*}
    \centering
\caption{Quantitative results on ShapeNet, Thingi10K~\cite{Thingi10K}, and Objaverse. The best \textbf{scores} are emphasized in bold. The second \underline{score} is emphasized in lines. * represents the resolution is 512.}
\begin{tabular}{c c c c c c c}
\toprule
\textbf{ Dataset} & \textbf{Method} & \textbf{HD} $\downarrow$ & $\mathrm{CD}_{L 1} \downarrow$ & $\mathrm{CD}_{L 2}\left(\times 10^3\right) \downarrow$ & $\mathrm{NC} \uparrow$ & F1 $\uparrow$  \\
\midrule
\multirow{5}{*}{ShapeNet} & MeshAnythingV2* & 0.078 & 0.009 & 0.640 & 0.911 & 0.652 \\
& BPT* & \textbf{0.017} & \textbf{0.003} & \textbf{0.012} & \underline{0.962} & \underline{0.875}  \\
& TreeMeshGPT* & 0.161 & 0.034 & 5.430 & 0.841 & 0.556\\
& DeepMesh* & 0.037 & 0.004 & 0.060 & \textbf{0.967} & 0.791  \\
& TSSR (Our) & \underline{0.050}& \underline{0.013} & \underline{0.035} & 0.954 & \textbf{0.976} \\
\midrule
\multirow{5}{*}{Thingi10K} & MeshAnythingV2* & 0.167 & \textbf{0.021} & \underline{2.492} & 0.842 & 0.358\\
& BPT* & \underline{0.157} & 0.035 & 7.771 & \underline{0.875} & \underline{0.496} \\
& TreeMeshGPT* & 0.233 & 0.060 & 18.086 & 0.788 & 0.387 \\
& DeepMesh* & 0.165 & \underline{0.026} & 3.331 & 0.853 & 0.321 \\
& TSSR (Our) & \textbf{0.106} & 0.032 & \textbf{1.183}& \textbf{0.969} & \textbf{0.775}\\
\midrule
\multirow{5}{*}{Objaverse} & MeshAnythingV2* & 0.118 & \underline{0.015} & \underline{1.213} & 0.859 & 0.430 \\
& BPT* & 0.151 & 0.034 & 7.016 & 0.846 & \underline{0.502} \\
& TreeMeshGPT* & 0.237 & 0.057 & 10.507 & 0.784 & 0.308 \\
& DeepMesh* & \underline{0.111} & 0.016 & 1.712 & \underline{0.866} & 0.471 \\
& TSSR (Our) & \textbf{0.101} & \textbf{0.014} & \textbf{0.863} & \textbf{0.959} & \textbf{0.744} \\
\bottomrule
\end{tabular}
    \label{tab:experiment}
\end{table*}

\begin{table*}[h]
    \centering
\caption{Ablation study of the face counts. S1-S3 represent the different strategies to perturb face count.}
\begin{tabular}{c c c c c c c}
\toprule
\textbf{ Dataset} & \textbf{Strategy} & \textbf{HD} $\downarrow$ & $\mathrm{CD}_{L 1} \downarrow$ & $\mathrm{CD}_{L 2}\left(\times 10^3\right) \downarrow$ & $\mathrm{NC} \uparrow$ & F1 $\uparrow$  \\
\midrule
\multirow{3}{*}{ShapeNet} & S1 & 0.050 & 0.013 & 0.035 & 0.954& 0.976 \\
& S2 & 0.054 & 0.012 & 0.347 & 0.841 & 0.970  \\
& S3 & 0.161 & 0.034 & 5.430 & 0.841 & 0.959\\
\midrule
\multirow{3}{*}{Thingi10K}& S1 & 0.106 & 0.032 & 1.183 & 0.969 & 0.775\\
& S2 & 0.124 & 0.032 & 1.240 & 0.915 & 0.731 \\
& S3 & 0.233 & 0.060 & 18.086 & 0.788 & 0.611 \\
\midrule
\multirow{3}{*}{Objaverse} & S1 & 0.101 & 0.014 & 0.863 & 0.959 & 0.744 \\
& S2 & 0.156 & 0.034 & 1.878 & 0.890 & 0.731 \\
& S3 & 0.237 & 0.057 & 10.507 & 0.784 & 0.641 \\
\bottomrule
\end{tabular}
    \label{tab:face_count}
\end{table*}
We report the comprehensive quantitative results shown in Table~\ref{tab:experiment}. TSSR achieves competitive results as an AR-like method. Concretely, in three datasets, TSSR achieves the best performance on all metrics. To further confirm the quality of the meshes, we present additional qualitative results in Fig.~\ref{tab:cost}. It can be found that TSSR can generate artist-like meshes without the topology errors.

Meanwhile, thanks to parallel token generation, the generated mesh will be more complete than the ARs. Since some meshes generated by ARs will lose some necessary faces. In this way, the experimental results prove the validity of TSSR.

\textbf{Computation cost.} To demonstrate the efficiency of the parallel token generation, we compare the computation cost of TSSR for generating 2k faces meshes with BPT and DeepMesh, shown in Table~\ref{tab:cost}.
\begin{table}
    \centering
\caption{Ablation study about computation cost on a single H20.}
\begin{tabular}{c c}
\toprule
\textbf{Method} & \textbf{Times (s)}  \\
\midrule
DeepMesh& 73 \\
BPT& 79\\
TSSR (Our) & \textbf{51} \\
\bottomrule
\end{tabular}
    \label{tab:cost}
\end{table}

\subsection{Ablation Study}
\textbf{Components on Backbone.} To illustrate the influence of each improvement for the improved Hourglass transformer, we report the ablation study shown in Table~\ref{tab:backbone}. It can be observed that adding cross-attention and multi-level RoPE can improve the quality of generated text.
\begin{table}
    \centering
\caption{Ablation study about components in Hourglass on ShapeNet. N/A indicates that the model cannot converge within the same number of training steps.}
\begin{tabular}{c c | c c }
\toprule
\textbf{Cross Attention} & \textbf{RoPE} & \textbf{HD} $\downarrow$ & $\mathrm{CD}_{L 1} \downarrow$   \\
\midrule
& & N/A & N/A \\
$\surd$ & & 0.1338& 0.025 \\
$\surd$ & $\surd$ & 0.050 & 0.013 \\
\bottomrule
\end{tabular}
    \label{tab:backbone}
\end{table}
\begin{table}
    \centering
\caption{Ablation study about connection loss on ShapeNet. W and W/O represent whether the connection loss is used during training, respectively. }
\begin{tabular}{c | c c }
\toprule
\text{Connection Loss} & \textbf{HD} $\downarrow$ & $\mathrm{CD}_{L 1} \downarrow$   \\
\midrule
W& 0.122 & 0.024\\
W/O& 0.050 & 0.013 \\
\bottomrule
\end{tabular}
    \label{tab:connection_loss}
\end{table}
\begin{figure}
    \centering
    \includegraphics[width=0.9\linewidth]{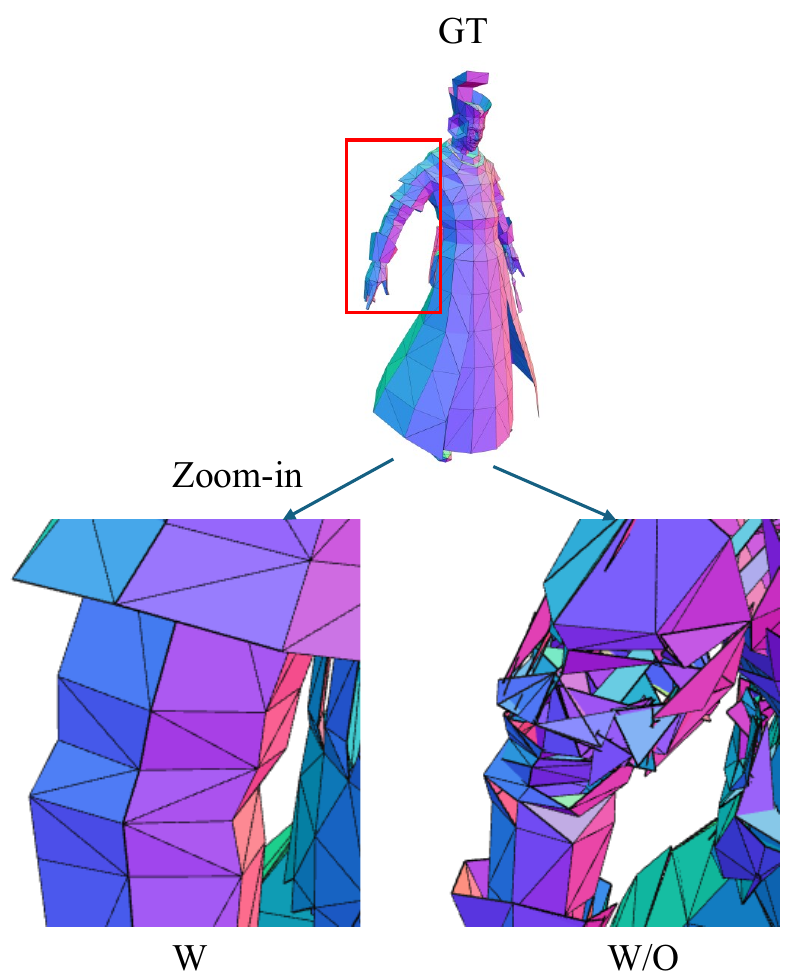}
    \caption{Qualitative results about the ablation study of connection loss. }
    \label{fig:connection_loss}
\end{figure}
\textbf{Ablation study about connection loss}. To illustrate the importance of connection loss, we present an ablation study in Table~\ref{tab:connection_loss} and related qualitative findings in Fig.~\ref{fig:connection_loss}. It can demonstrate that the connection can enhance the topology quality. After canceling, the generated mesh will contain some fragile triangles. These triangles are in the correct place according to the cloud point condition, but they generate an invalid mesh.

\textbf{Face counts.} We show a powerful ability of TSSR that can precisely control the face counts in Table~\ref{tab:face_count}. We implement three strategies: S1) Using the GT face count. S2) Introducing bias (the face counts will randomly increase/decrease among 100 faces). S3) Maximize bias (the face counts will keep a constant value, i.e., 2k). We also report detailed qualitative results in the supplementary material. Interestingly, TSSR can generate topologies that are entirely different yet artist-like via different face counts, demonstrating its potential.

\section{Conclusion}
In this paper, we propose TSSR, the DDM-based parallel token generation for direct mesh generation tasks. The motivation of TSSR is to maintain high token prediction accuracy after enabling token parallel generation. TSSR is composed of three novel components: 1) the decouple training and hybrid inference process. 2) An improved Hourglass architecture. 3) A novel connection. Unify them into one, TSSR successfully generates artist-like meshes via parallel token generation. Experimental results demonstrate that TSSR can generate 10K face meshes up to 1024 resolution.

\textbf{Limitation.} Similar to the DDM-based LLM, TSSR also suffers from inconvenient training. For example, we need to pad the sequence length to accommodate the large batch size, which will inevitably incur some computational cost. Such a problem could be resolved via infilling, which remains future work. 
{
    \small
    \bibliographystyle{ieeenat_fullname}
    \bibliography{main}

\begin{thebibliography}{29}
\providecommand{\natexlab}[1]{#1}
\providecommand{\url}[1]{\texttt{#1}}
\expandafter\ifx\csname urlstyle\endcsname\relax
  \providecommand{\doi}[1]{doi: #1}\else
  \providecommand{\doi}{doi: \begingroup \urlstyle{rm}\Url}\fi

\bibitem[Alliegro et~al.(2023)Alliegro, Siddiqui, Tommasi, and Nießner]{polydiff}
Antonio Alliegro, Yawar Siddiqui, Tatiana Tommasi, and Matthias Nießner.
\newblock Polydiff: Generating 3d polygonal meshes with diffusion models, 2023.

\bibitem[Chang et~al.(2015)Chang, Funkhouser, Guibas, Hanrahan, Huang, Li, Savarese, Savva, Song, Su, Xiao, Yi, and Yu]{shapenet}
Angel~X. Chang, Thomas Funkhouser, Leonidas Guibas, Pat Hanrahan, Qixing Huang, Zimo Li, Silvio Savarese, Manolis Savva, Shuran Song, Hao Su, Jianxiong Xiao, Li Yi, and Fisher Yu.
\newblock Shapenet: An information-rich 3d model repository, 2015.

\bibitem[Chen et~al.(2024{\natexlab{a}})Chen, Chen, Pang, Zeng, Cheng, Fu, Yin, Wang, Yu, Yu, Fu, and Chen]{MeshXL}
Sijin Chen, Xin Chen, Anqi Pang, Xianfang Zeng, Wei Cheng, Yijun Fu, Fukun Yin, Billzb Wang, Jingyi Yu, Gang Yu, Bin Fu, and Tao Chen.
\newblock Meshxl: Neural coordinate field for generative 3d foundation models.
\newblock In \emph{Advances in Neural Information Processing Systems 38: Annual Conference on Neural Information Processing Systems 2024, NeurIPS 2024, Vancouver, BC, Canada, December 10 - 15, 2024}, 2024{\natexlab{a}}.

\bibitem[Chen et~al.(2024{\natexlab{b}})Chen, He, Huang, Ye, Chen, Tang, Chen, Cai, Yang, Yu, Lin, and Zhang]{meshanything}
Yiwen Chen, Tong He, Di Huang, Weicai Ye, Sijin Chen, Jiaxiang Tang, Xin Chen, Zhongang Cai, Lei Yang, Gang Yu, Guosheng Lin, and Chi Zhang.
\newblock Meshanything: Artist-created mesh generation with autoregressive transformers, 2024{\natexlab{b}}.

\bibitem[Chen et~al.(2024{\natexlab{c}})Chen, Wang, Luo, Wang, Chen, Zhu, Zhang, and Lin]{meshanythingv2}
Yiwen Chen, Yikai Wang, Yihao Luo, Zhengyi Wang, Zilong Chen, Jun Zhu, Chi Zhang, and Guosheng Lin.
\newblock Meshanything v2: Artist-created mesh generation with adjacent mesh tokenization, 2024{\natexlab{c}}.

\bibitem[Deitke et~al.(2023)Deitke, Liu, Wallingford, Ngo, Michel, Kusupati, Fan, Laforte, Voleti, Gadre, VanderBilt, Kembhavi, Vondrick, Gkioxari, Ehsani, Schmidt, and Farhadi]{objxl}
Matt Deitke, Ruoshi Liu, Matthew Wallingford, Huong Ngo, Oscar Michel, Aditya Kusupati, Alan Fan, Christian Laforte, Vikram Voleti, Samir~Yitzhak Gadre, Eli VanderBilt, Aniruddha Kembhavi, Carl Vondrick, Georgia Gkioxari, Kiana Ehsani, Ludwig Schmidt, and Ali Farhadi.
\newblock Objaverse-xl: A universe of 10m+ 3d objects, 2023.

\bibitem[Gat et~al.(2024)Gat, Remez, Shaul, Kreuk, Chen, Synnaeve, Adi, and Lipman]{dfm}
Itai Gat, Tal Remez, Neta Shaul, Felix Kreuk, Ricky T.~Q. Chen, Gabriel Synnaeve, Yossi Adi, and Yaron Lipman.
\newblock Discrete flow matching, 2024.

\bibitem[Hao et~al.(2024)Hao, Romero, Lin, and Liu]{meshtron}
Zekun Hao, David~W. Romero, Tsung-Yi Lin, and Ming-Yu Liu.
\newblock Meshtron: High-fidelity, artist-like 3d mesh generation at scale, 2024.

\bibitem[He et~al.(2025)He, Chen, Huang, Liu, Huang, Ouyang, Yuan, and Li]{meshcraft}
Xianglong He, Junyi Chen, Di Huang, Zexiang Liu, Xiaoshui Huang, Wanli Ouyang, Chun Yuan, and Yangguang Li.
\newblock Meshcraft: Exploring efficient and controllable mesh generation with flow-based dits, 2025.

\bibitem[Hunyuan3D et~al.(2025)Hunyuan3D, Yang, Yang, Feng, Huang, Zhang, He, Luo, Liu, Zhao, Lin, Lai, Yang, Shi, Zhao, Zhang, Yan, Wang, Liu, Zhang, Chen, Dong, Jia, Cai, Yu, Tang, Guo, Yu, Zhang, Ye, He, Wu, Wei, Zhang, Tan, Sun, Niu, Huang, Zheng, Liu, Chen, Yuan, Yang, Liu, Zhu, Chen, Liu, Wang, Liu, Linus, Jiang, Huang, and Guo]{hunyuan2.1}
Team Hunyuan3D, Shuhui Yang, Mingxin Yang, Yifei Feng, Xin Huang, Sheng Zhang, Zebin He, Di Luo, Haolin Liu, Yunfei Zhao, Qingxiang Lin, Zeqiang Lai, Xianghui Yang, Huiwen Shi, Zibo Zhao, Bowen Zhang, Hongyu Yan, Lifu Wang, Sicong Liu, Jihong Zhang, Meng Chen, Liang Dong, Yiwen Jia, Yulin Cai, Jiaao Yu, Yixuan Tang, Dongyuan Guo, Junlin Yu, Hao Zhang, Zheng Ye, Peng He, Runzhou Wu, Shida Wei, Chao Zhang, Yonghao Tan, Yifu Sun, Lin Niu, Shirui Huang, Bojian Zheng, Shu Liu, Shilin Chen, Xiang Yuan, Xiaofeng Yang, Kai Liu, Jianchen Zhu, Peng Chen, Tian Liu, Di Wang, Yuhong Liu, Linus, Jie Jiang, Jingwei Huang, and Chunchao Guo.
\newblock Hunyuan3d 2.1: From images to high-fidelity 3d assets with production-ready pbr material, 2025.

\bibitem[Lai et~al.(2025)Lai, Zhao, Liu, Zhao, Lin, Shi, Yang, Yang, Yang, Feng, Zhang, Huang, Luo, Yang, Yang, Wang, Liu, Tang, Cai, He, Liu, Liu, Jiang, Linus, Huang, and Guo]{hunyuan2.5}
Zeqiang Lai, Yunfei Zhao, Haolin Liu, Zibo Zhao, Qingxiang Lin, Huiwen Shi, Xianghui Yang, Mingxin Yang, Shuhui Yang, Yifei Feng, Sheng Zhang, Xin Huang, Di Luo, Fan Yang, Fang Yang, Lifu Wang, Sicong Liu, Yixuan Tang, Yulin Cai, Zebin He, Tian Liu, Yuhong Liu, Jie Jiang, Linus, Jingwei Huang, and Chunchao Guo.
\newblock Hunyuan3d 2.5: Towards high-fidelity 3d assets generation with ultimate details, 2025.

\bibitem[Lionar et~al.(2025)Lionar, Liang, and Lee]{TreeMeshGPT}
Stefan Lionar, Jiabin Liang, and Gim~Hee Lee.
\newblock Treemeshgpt: Artistic mesh generation with autoregressive tree sequencing.
\newblock In \emph{{IEEE/CVF} Conference on Computer Vision and Pattern Recognition, {CVPR} 2025, Nashville, TN, USA, June 11-15, 2025}, pages 26608--26617, 2025.

\bibitem[Liu et~al.(2022)Liu, Gong, and Liu]{fm}
Xingchao Liu, Chengyue Gong, and Qiang Liu.
\newblock Flow straight and fast: Learning to generate and transfer data with rectified flow, 2022.

\bibitem[Long et~al.(2024)Long, Guo, Lin, Liu, Dou, Liu, Ma, Zhang, Habermann, Theobalt, and Wang]{wonder3d}
Xiaoxiao Long, Yuan{-}Chen Guo, Cheng Lin, Yuan Liu, Zhiyang Dou, Lingjie Liu, Yuexin Ma, Song{-}Hai Zhang, Marc Habermann, Christian Theobalt, and Wenping Wang.
\newblock Wonder3d: Single image to 3d using cross-domain diffusion.
\newblock In \emph{{IEEE/CVF} Conference on Computer Vision and Pattern Recognition, {CVPR} 2024, Seattle, WA, USA, June 16-22, 2024}, pages 9970--9980, 2024.

\bibitem[Lorensen and Cline(1987)]{marchingcubes}
William~E. Lorensen and Harvey~E. Cline.
\newblock Marching cubes: {A} high resolution 3d surface construction algorithm.
\newblock In \emph{Proceedings of the 14th Annual Conference on Computer Graphics and Interactive Techniques, {SIGGRAPH} 1987, Anaheim, California, USA, July 27-31, 1987}, pages 163--169, 1987.

\bibitem[Loshchilov and Hutter(2019)]{adamw}
Ilya Loshchilov and Frank Hutter.
\newblock Decoupled weight decay regularization, 2019.

\bibitem[Mildenhall et~al.(2020)Mildenhall, Srinivasan, Tancik, Barron, Ramamoorthi, and Ng]{NeRF}
Ben Mildenhall, Pratul~P. Srinivasan, Matthew Tancik, Jonathan~T. Barron, Ravi Ramamoorthi, and Ren Ng.
\newblock Nerf: Representing scenes as neural radiance fields for view synthesis, 2020.

\bibitem[Nie et~al.(2025)Nie, Zhu, You, Zhang, Ou, Hu, Zhou, Lin, Wen, and Li]{llada}
Shen Nie, Fengqi Zhu, Zebin You, Xiaolu Zhang, Jingyang Ou, Jun Hu, Jun Zhou, Yankai Lin, Ji-Rong Wen, and Chongxuan Li.
\newblock Large language diffusion models, 2025.

\bibitem[Shaul et~al.(2025)Shaul, Gat, Havasi, Severo, Sriram, Holderrieth, Karrer, Lipman, and Chen]{fmgdp}
Neta Shaul, Itai Gat, Marton Havasi, Daniel Severo, Anuroop Sriram, Peter Holderrieth, Brian Karrer, Yaron Lipman, and Ricky T.~Q. Chen.
\newblock Flow matching with general discrete paths: {A} kinetic-optimal perspective.
\newblock In \emph{The Thirteenth International Conference on Learning Representations, {ICLR} 2025, Singapore, April 24-28, 2025}. OpenReview.net, 2025.

\bibitem[Siddiqui et~al.(2024)Siddiqui, Alliegro, Artemov, Tommasi, Sirigatti, Rosov, Dai, and Nie{\ss}ner]{MeshGPT}
Yawar Siddiqui, Antonio Alliegro, Alexey Artemov, Tatiana Tommasi, Daniele Sirigatti, Vladislav Rosov, Angela Dai, and Matthias Nie{\ss}ner.
\newblock Meshgpt: Generating triangle meshes with decoder-only transformers.
\newblock In \emph{{IEEE/CVF} Conference on Computer Vision and Pattern Recognition, {CVPR} 2024, Seattle, WA, USA, June 16-22, 2024}, pages 19615--19625, 2024.

\bibitem[Song et~al.(2025)Song, Zhang, Luo, Gao, Xia, Luo, Li, Yang, Yu, Qu, Fu, Su, Zhang, Huang, Wang, Yan, Jia, Liu, Ma, Zhang, Wu, and Zhou]{seeddiffusion}
Yuxuan Song, Zheng Zhang, Cheng Luo, Pengyang Gao, Fan Xia, Hao Luo, Zheng Li, Yuehang Yang, Hongli Yu, Xingwei Qu, Yuwei Fu, Jing Su, Ge Zhang, Wenhao Huang, Mingxuan Wang, Lin Yan, Xiaoying Jia, Jingjing Liu, Wei-Ying Ma, Ya-Qin Zhang, Yonghui Wu, and Hao Zhou.
\newblock Seed diffusion: A large-scale diffusion language model with high-speed inference, 2025.

\bibitem[Tang et~al.()Tang, Li, Hao, Liu, Zeng, Liu, and Zhang]{edgerunner}
Jiaxiang Tang, Zhaoshuo Li, Zekun Hao, Xian Liu, Gang Zeng, Ming{-}Yu Liu, and Qinsheng Zhang.
\newblock Edgerunner: Auto-regressive auto-encoder for artistic mesh generation.
\newblock In \emph{The Thirteenth International Conference on Learning Representations, {ICLR} 2025, Singapore, April 24-28, 2025}.

\bibitem[Wang et~al.(2025)Wang, Lai, Li, Zhang, Sun, Kang, Wu, Li, and Luo]{FUDOKI}
Jin Wang, Yao Lai, Aoxue Li, Shifeng Zhang, Jiacheng Sun, Ning Kang, Chengyue Wu, Zhenguo Li, and Ping Luo.
\newblock Fudoki: Discrete flow-based unified understanding and generation via kinetic-optimal velocities, 2025.

\bibitem[Weng et~al.(2024)Weng, Zhao, Lei, Yang, Liu, Lai, Chen, Liu, Jiang, Guo, Zhang, Gao, and Chen]{bpt}
Haohan Weng, Zibo Zhao, Biwen Lei, Xianghui Yang, Jian Liu, Zeqiang Lai, Zhuo Chen, Yuhong Liu, Jie Jiang, Chunchao Guo, Tong Zhang, Shenghua Gao, and C.~L.~Philip Chen.
\newblock Scaling mesh generation via compressive tokenization, 2024.

\bibitem[Xiang et~al.(2025)Xiang, Lv, Xu, Deng, Wang, Zhang, Chen, Tong, and Yang]{TRELLIS}
Jianfeng Xiang, Zelong Lv, Sicheng Xu, Yu Deng, Ruicheng Wang, Bowen Zhang, Dong Chen, Xin Tong, and Jiaolong Yang.
\newblock Structured 3d latents for scalable and versatile 3d generation.
\newblock In \emph{{IEEE/CVF} Conference on Computer Vision and Pattern Recognition, {CVPR} 2025, Nashville, TN, USA, June 11-15, 2025}, pages 21469--21480, 2025.

\bibitem[Xu et~al.(2025)Xu, Xue, Dong, Wan, Zhu, Li, Dou, Lin, Xin, Liu, Wang, and Komura]{MeshMosaic}
Rui Xu, Tianyang Xue, Qiujie Dong, Le Wan, Zhe Zhu, Peng Li, Zhiyang Dou, Cheng Lin, Shiqing Xin, Yuan Liu, Wenping Wang, and Taku Komura.
\newblock Meshmosaic: Scaling artist mesh generation via local-to-global assembly, 2025.

\bibitem[Zhao et~al.(2025{\natexlab{a}})Zhao, Ye, Wang, Liu, Chen, Wang, and Zhu]{deepmesh}
Ruowen Zhao, Junliang Ye, Zhengyi Wang, Guangce Liu, Yiwen Chen, Yikai Wang, and Jun Zhu.
\newblock Deepmesh: Auto-regressive artist-mesh creation with reinforcement learning, 2025{\natexlab{a}}.

\bibitem[Zhao et~al.(2025{\natexlab{b}})Zhao, Lai, Lin, Zhao, Liu, Yang, Feng, Yang, Zhang, Yang, Shi, Liu, Wu, Lian, Yang, Tang, He, Wang, Liu, Zuo, Chen, Lei, Weng, Xu, Zhu, Liu, Xu, Hu, Yang, Zhang, Liu, Huang, Wang, Zhang, Chen, Dong, Jia, Cai, Yu, Tang, Zhang, Ye, He, Wu, Zhang, Tan, Xiao, Tao, Zhu, Xue, Liu, Zhao, Wu, Hu, Qin, Peng, Li, Chen, Zhang, Niu, Wang, Wang, Kuang, Fan, Zheng, Zhuang, He, Liu, Yang, Wang, Liu, Jiang, Huang, and Guo]{hunyuan2.0}
Zibo Zhao, Zeqiang Lai, Qingxiang Lin, Yunfei Zhao, Haolin Liu, Shuhui Yang, Yifei Feng, Mingxin Yang, Sheng Zhang, Xianghui Yang, Huiwen Shi, Sicong Liu, Junta Wu, Yihang Lian, Fan Yang, Ruining Tang, Zebin He, Xinzhou Wang, Jian Liu, Xuhui Zuo, Zhuo Chen, Biwen Lei, Haohan Weng, Jing Xu, Yiling Zhu, Xinhai Liu, Lixin Xu, Changrong Hu, Shaoxiong Yang, Song Zhang, Yang Liu, Tianyu Huang, Lifu Wang, Jihong Zhang, Meng Chen, Liang Dong, Yiwen Jia, Yulin Cai, Jiaao Yu, Yixuan Tang, Hao Zhang, Zheng Ye, Peng He, Runzhou Wu, Chao Zhang, Yonghao Tan, Jie Xiao, Yangyu Tao, Jianchen Zhu, Jinbao Xue, Kai Liu, Chongqing Zhao, Xinming Wu, Zhichao Hu, Lei Qin, Jianbing Peng, Zhan Li, Minghui Chen, Xipeng Zhang, Lin Niu, Paige Wang, Yingkai Wang, Haozhao Kuang, Zhongyi Fan, Xu Zheng, Weihao Zhuang, YingPing He, Tian Liu, Yong Yang, Di Wang, Yuhong Liu, Jie Jiang, Jingwei Huang, and Chunchao Guo.
\newblock Hunyuan3d 2.0: Scaling diffusion models for high resolution textured 3d assets generation, 2025{\natexlab{b}}.

\bibitem[Zhou and Jacobson(2016)]{Thingi10K}
Qingnan Zhou and Alec Jacobson.
\newblock Thingi10k: A dataset of 10,000 3d-printing models.
\newblock \emph{arXiv preprint arXiv:1605.04797}, 2016.

\end{thebibliography}
}

\end{document}